\documentclass[runningheads]{llncs}
\usepackage{graphicx}

\usepackage{caption}
\usepackage{subcaption}

\begin{document}
\title{Using Explainable Boosting Machine to Compare Idiographic and Nomothetic Approaches for Ecological Momentary Assessment Data}
\titlerunning{EBMs to Compare Idiographic and Nomothetic Approaches}
%
\author{Mandani Ntekouli\inst{1} \and
Gerasimos Spanakis\inst{1} \and
Lourens Waldorp \inst{2} \and
Anne Roefs\inst{3}}
\authorrunning{M. Ntekouli et al.}
%
\institute{Department of Data Science and Knowledge Engineering,
Maastricht University, Maastricht, Netherlands \and
Department of Psychological Methods, University of Amsterdam, Amsterdam, Netherlands
\and
Faculty of Psychology and Neuroscience, Maastricht
University, Maastricht, Netherlands
}
\maketitle              
\begin{abstract}

Previous research on EMA data of mental disorders was mainly focused on multivariate regression-based approaches modeling each individual separately. This paper goes a step further towards exploring the use of non-linear interpretable machine learning (ML) models in classification problems. ML models can enhance the ability to accurately predict the occurrence of different behaviors by recognizing complicated patterns between variables in data.
To evaluate this, the performance of various ensembles of trees are compared to linear models using imbalanced synthetic and real-world datasets. After examining the distributions of AUC scores in all cases, non-linear models appear to be superior to baseline linear models.
Moreover, apart from personalized approaches, group-level prediction models are also likely to offer an enhanced performance. According to this, two different nomothetic approaches to integrate data of more than one individuals are examined, one using directly all data during training and one based on knowledge distillation. 
Interestingly, it is observed that in one of the two real-world datasets, knowledge distillation method achieves improved AUC scores (mean relative change of +17\% compared to personalized) showing how it can benefit EMA data classification and performance.

\end{abstract}
\section{Introduction}

In the last few years, there has been a renewed research interest in the areas of psychology and psychiatry that has been particularly sparked by recent technological and methodological developments for collecting time-intensive, repeated, intra-individual measurements through Ecological Momentary Assessment (EMA) studies \cite{fried17}, \cite{psy_review}, \cite{ema1}, \cite{ema3}. EMA offers the opportunity to capture relevant information about patients' evolution of their mental condition, symptoms and experiences, in real-time and in context of their everyday life. This way, a large amount of personalized data has become available, providing the means for further exploring mental disorders \cite{tvar}. Consequently, there has been an urgent need for developing statistical methods to model psychological behaviour \cite{gvar}. Some practical applications of such models could be to predict illness course, determine treatment response or develop tailored psychiatric interventions \cite{predict}.

Based on literature, EMA time-series data have been mostly studied in a multivariate regression-based approach \cite{gvar}, \cite{var1}. More specifically, the most popular class of time-series models is the Vector Autoregressive (VAR) model with a goal to estimate the dynamical interactions between all the measured variables (i.e., network structures) \cite{network}. 
However, the fact that these models can only estimate linear statistical relationships can be a significant issue for mental disorders, where the involved interactions are likely to be quite complex. When many symptoms or variables are involved in the course, these are more prone to interact in a non-linear fashion with each other. Thus, linear models seem insufficient to uncover the possible non-linear interactions and describe precisely the real complex nature of mental disorders.

A promising approach that can learn such complex and higher-order interactions of symptoms is using non-linear machine learning (ML) models \cite{esm-ml}. ML models can enhance the ability to accurately predict the occurrence of different behaviors by recognizing complicated patterns or relations between variables in existing data. 

This work focuses on two research objectives, examining the idiographic (or personalized) and nomothetic (or group-based) predictive approach, respectively. First, according to the idiographic approach, personalized models are typically applied, as there are possibly different underlying mechanisms that drive a future behavior in each individual. Thus, different non-linear interpretable models are evaluated in terms of performance to test whether they are superior to baseline linear models. Second, we should acknowledge the possibility that shared influences among different individuals may provide a complementary predictive utility. Therefore, prediction models are applied in a nomothetic (group-based) approach showing that integrating data of more than one individual in a single model could also accurately predict future outcomes at a person level \cite{nomo}. 

\section{Methodology}

\subsection{Idiographic (person-specific) approach}
Based on the fact that mental disorders can be modeled as a complex system, we assume that illness course and behaviors differ remarkably across individuals. Most individuals suffering from the same disorder are likely to exhibit different symptoms, so different mechanisms possibly influence and drive a future behavior. Therefore, it is proposed that each individual should be examined separately using personalized prediction models \cite{fried17}. 

Starting from the widely used linear models, a natural extension of these is the more flexible Generalized Additive Models (GAMs) \cite{gam}. The main concept of GAMs remains the same as of the linear ones, expecting for the outcome to be an additive model of feature effects, but relaxing the restriction of the linear relationship.
It allows the use of arbitrary functions for representing the features’ effects. Subsequently, more flexible, non-linear feature functions can be incorporated. 
These functions can be based on regression spline models and tree-based models such as single trees or ensembles of bagged trees, boosted trees and boosted-bagged trees.

However, there is still a significant gap between the flexible GAMs and full-complexity models, such as ensembles of trees, regarding accuracy \cite{gam}. The main reason of this limitation is that GAMs take into account only univariate terms and not any interaction between features (variables).
To deal with this drawback, a more advanced method was developed, called Generalized Additive
Models plus Interactions ($GA^2Ms$), which incorporates pairwise interactions between features \cite{gam2}. The model is described in the following form:

\begin{equation}
    g(y) = \sum_i f_i(x_i) + \sum_{i\neq j} f_{ij}(x_i, x_j)
\end{equation}
 
where $f$ are the feature functions of features $x$ and $g$ is the link function (eg. identity or logistic) of the predicted outcome $y$.
This model can still be interpretable, using heat maps for representing the pairwise features’ interactions, as well as accurate, reaching the performance of the state-of-the-art ML models.

In this work, a fast implementation of the $GA^2Ms$ algorithm was used, called Explainable Boosting Machine models (EBMs), which is part of the Microsoft’s open-source Python package InterpretML \cite{ebm}. 
The EBMs’ learning process makes use of gradient boosting with shallow regression tree ensembles. At each boosting round, a tree is built on a single feature and its residuals are used for training the tree of the following feature. This is repeated for all different features. After several boosting rounds, each feature’s trees of all rounds can be combined, leading to tree ensembles as the final features’ representation. On top of this, functions for pairwise features’ interactions can be additionally incorporated. The FAST method is used to detect and rank features’ interactions in order to keep the most significant ones, without the expense of checking all possible combinations \cite{gam2}. Again, the same training process is performed for the specified pairs. 

Because EBMs is a relatively novel method, its performance is evaluated by comparing it to other full-complexity ML models, such as XGBoost, Gradient Boosting Trees (GradBoost) and Random Forest (RF). Afterwards, non-linear models are also compared to linear models, such as Logistic Regression (LogReg) and Support Vector Machines (SVM), using a linear kernel.

\subsection{Nomothetic (group-level) Approaches}
\label{method}
Although personalized models are mostly applied, commonalities among different participants may provide complementary predictive utility. Thus, population-level prediction studies are also likely to offer an enhanced performance. Especially, in case of more advanced ML models, incorporating more data could be of more help, compared to the traditional linear models. This approach could have a clear advantage to uncover potential complex hidden relationships between variables. 

The most common way of integrating data of more than one individual in a model is to concatenate the data of all individuals together in a single dataset. The augmented dataset is then used to construct a population-based model. Such models produce generalizable predications that can be relevant to a wider range of individuals. For example, a population-based model can be applied to new individuals who have not been included in the training of the model. An additional benefit would be that it can be applied to individuals that cannot be run in a personalized way due to the lack of the necessary amount of training samples (time-points) or samples of the minority class.

The second proposed approach is based on the Knowledge Distillation (also known as teacher-student) method \cite{dist}. In this case, information from a larger (teacher) model is used in a smaller (student) model. 
The original concept of Knowledge Distillation was created with the goal to fill the gap between expressive power and learnability in Neural Networks (NNs). This is achieved by training a small NN after incorporating additional information from a larger and more complex NN. However, the aforementioned gap does not only exist in NNs but also in other machine learning methods like the tree-based models described above \cite{dist2}. So, the distillation method using information extracted from larger models can be further exploited in non-NN models.

\begin{figure}[!ht]
    \centering
    \includegraphics[width = 0.6\textwidth]{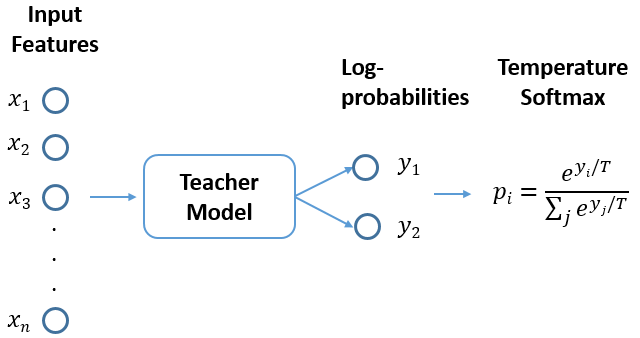}
    \caption{The proposed Knowledge Distillation method: After each sample $(x1,x2,..,xn)$ is input to the teacher model, the extracted log-probabilies $y_i$ are used to the temperature softmax function. The produced $p_1$ or $p_2$ are the labels for the student models.}
    \label{dist_arch}
\end{figure}

Inspired by this, the proposed Knowledge Distillation method in our case is illustrated in Fig.~\ref{dist_arch}. First, the teacher model is trained on data from all participants and then information derived from this model is used for training personalized student models for each individual separately. Additional information is gained through the smoothened probabilities $p_i$ (soft labels), which come from the pre-trained teacher model. The log-probabilities $y_i$ of the training samples are softened using a temperature softmax function.
The temperature hyperparameter $T$ plays an important role in smoothing the distribution of the outputs, that is necessary to prevent the case of having an one-hot vector as the result of a typical softmax function. Then, the smoothened outputs are used as labels for the personalized student models. Compared to the conventional personalized training that uses hard labels, distillation can provide additional useful information with an aim to improve the personalized models.

\section{Experimental Setup}

\subsection{EMA Datasets}
\label{dataset}

EMA data is organized in a hierarchical structure for each individual, where observations are collected multiple times a day for a pre-defined period of several weeks. The total number of observations as well as the collection period can be different among individuals because some may experience difficulties in following the schedule of the surveys. All datasets' characteristics used in this paper are briefly reported in Table \ref{tab:data} and more extensively presented as follows.

\begin{table*}[t]
\centering
\caption{Characteristics of the examined datasets. For imbalance ratio and training/test sets, the mean and standard deviation values of all individuals are presented, after pre-processing.}
\scalebox{0.825}{
\begin{tabular}{l c c | c c c}
\hline
   \textbf{Dataset}  & \textbf{\#Participants} & \textbf{\#Features}  &
   \textbf{Imbalance Ratio} &
   \textbf{\#Training Samples} & \textbf{\#Test Samples}\\
   \hline
   Synthetic & 20, 50, 100 & 25, 60  & $2.33$ & $35, 70, 210 $ & $15, 30, 90$\\
   Drink & 24 & 15  & $8.45 (5.45)$ & $72.83 (12.02)$ & $31.87 (5.24)$\\
    ThinkSlim2 & 57 & 43  & $5.82 (3.25)$ & $86 (38.72)$ & $37.51 (16.68)$\\
\end{tabular}
}
\label{tab:data}
\end{table*}

\subsubsection{Synthetic Datasets}

Due to lack of access to big EMA datasets, we follow a simple method for generating random EMA datasets. These datasets are designed to consist of the feature vectors and labels of each patient, aiming at a 2-class classification problem. It is also commonly noticed that medical-related EMA datasets, as well as the following examined real-world datasets, are characterized as imbalanced. This means that the majority of samples belongs only to one class, whereas much fewer to the other class. Thus, in this case, the ratio of samples assigned to the two classes is $0.7:0.3$ in the synthetic datasets as well. 

Furthermore, the datasets must be created in a way to be structurally similar to the real EMA data. First, these must incorporate multivariate ordinal and categorical variables. This is a challenging issue, especially in high dimensional datasets. The method of generating our feature vectors is based on sampling from a different random normal distribution for each one. These are afterwards transformed into ordinal features after deriving an equi-width histogram of the distribution, leading to a random number of 6 distinct values, or 2 as categorical variables are typically encoded to binary.

It is also often necessary to impose some flexibility on the data variables, such as noise. Noise can be added to both output labels and feature vectors. Here, a small amount of noise is introduced. More specifically, 20\% of the labels are randomly assigned to samples as well as the values of 20\% of features are randomly shuffled. Finally, regarding other characteristics of the synthetic datasets, such as the number of individuals, features and samples, a number of choices is evaluated.

\subsubsection{Dataset: Drink}
This first real-world dataset was obtained by a study described in \cite{alcohol}. It was a 2-week collection of data from 33 individuals through 8 daily mobile notifications/surveys. 
The captured variables included positive and negative emotions, drinking craving and expectancies, self-reported alcohol consumption, impulsivity, as well as social context. All these variables were measured on a scale from 0 to 100. Regarding the output variable, the aim of this prediction was the occurence or not of drinking events at the next time-point. So, a positive label was assigned to each sample when the number of alcoholic drinks at the next time-point was one or higher.

\subsubsection{Dataset: ThinkSlim2}
The second real-world dataset is larger and more challenging. It was obtained by a study described in more detail in \cite{thinkslim}, \cite{eat}.
This consisted of data collected from 134 overweight individuals multiple times a day (minimum 8) for 7 weeks (excluding the follow-up phase) via a mobile application. 
From all the measured variables, only some were selected based on the individuals’s compliance. The final variables included various positive and negative emotions, location, activity, social context and type of consumed food. The emotion-related variables were measured on a scale from 0 to 10. All other variables were categorical, including a set of predefined choices for each one. Regarding the output variable, the examined scenario was aiming at predicting the next healthy or unhealthy
eating event. So, a healthy or unhealthy label was assigned to each sample according to the type of food consumed at the next time-point. 

\subsection{Data Preparation}
\label{preparation}
For each dataset, each participant’s EMA data was prepared for analysis separately. These were assessed for the frequency of daily observations as well as the frequency and distribution of the outcome events. First, individuals having very few observations per day or in total were removed. The number of individuals retained was 31 and 76 for the Drink and ThinkSlim2 datasets, respectively. 

Additionally, because of the final goal to predict (or classify) the next time-point behavior, consecutive data points had to be collected. For example, for each data point, if the following one (collected within the next 2 hours) was absent then we could not retrieve its prediction target and eventually it was also considered as missing. That way, some individuals were found to have so few outcome events of the minority class that subsequent cross-validation steps could not be conducted. So, these participants were also excluded from the final dataset. As a result, the number of retained individuals were 24 for the Drink dataset having an average of 6.18 (std = 0.90) daily points and 57 for ThinkSlim2 with average 3.39 (std = 2.05) points.

As further seen in Table \ref{tab:data}, 
data points of each individual were split sequentially at fixed time intervals into two datasets, a training and a test set, containing the first 70\% and last 30\% of the data points, respectively. 

\subsection{Data Analysis}
\subsubsection{Idiographic Approach}
According to the idiographic approach, separate predictive models were applied to each individual, using various ML algorithms. A necessary step is hyperparameters’ tuning, which frequently has a big impact on model’s performance. In this paper, a time-series cross-validation method (a variation of KFold, returning first $k$ folds as training set and the $(k+1)$th fold as test set) was used to tune some of the main hyperparameters of the tree-based methods. All these combinations were exhaustively explored for each case using Grid Search and the one resulting to the best cross-validation score was retained for the following analysis. The metric score of interest was \emph{ROC\_AUC} (or any of the macro average scores), measuring the true-positive rate and false-positive rate for the model’s predictions using a set of different probability thresholds. AUC score was chosen for the prediction of both classes to be taken into account equally, regardless the number of samples these classes contained. In other words, the prediction of samples belonging to the majority class should not play a more important role than predicting samples of the minority class. 

\subsubsection{Nomothetic Approach}
According to the nomothetic approach, the two methods described in Section \ref{method} were investigated using Explainable Boosting Machine models (EBMs). EBMs were built using data of all individuals and then compared to the traditional personalized EBMs. In the first method, the training datasets of all individuals were concatenated in a population-level dataset, which was used to train an EBM. The number of interactions was fine-tuned to select the optimal value, as in the personalized models. The performance of this ``EBM\_all" model was evaluated separately on the testing set of each individual. The testing sets are kept the same as in the personalized approach.

In the second method, information obtained from the first method (teacher model) was further used in personalized EBMs. Each class' log-probabilities of the training samples were extracted and transformed to smoothed probabilities using a temperature softmax function, with the temperature value being selected from a range between $2$ and $200$. Thus, new datasets were created using the training samples of each individual and the extracted ``probabilities" as a target label, instead of the initial hard labels $(0, 1)$. These new datasets created for each individual were used to train the student models, which are EBMs regression models.

\section{Experimental Results}

\subsection{Synthetic Dataset}
\subsubsection{Idiographic Approach}
The initial step to evaluate the described methods was to create synthetic datasets. Using synthetic data, it is easier to understand the problem we have to solve and develop effective and efficient methods for that. To create the data, different values for the dataset’s parameters, such as number of subjects, features and samples, were chosen and investigated independently.

Synthetic datasets are first analyzed using a personalized approach. For each combination of the chosen parameters, personalized non-linear and linear models are applied to each individual of every dataset separately. After applying all personalized models, the mean and standard deviation values of the performance (AUC scores) of all created individuals are presented in Table \ref{tab:syn_idio}. It is clearly visible that the average AUC scores are greater when applying non-linear models. According to the extracted AUC results, EBMs models produce the best average scores in most of the datasets. However, even when RF or XGBoost show the best scores, their difference to EBMs is quite small. Moreover, EBMs achieved  more  accurate  performance when a large number of samples is used for training, such as 100 or 300.

\begin{table}[t]
\caption{Performance of personalized models (EBM, XGBoost, Gradient Boosting, RF, SVM and Logistic Regression): the mean and standard deviation of the AUC scores are given for all synthetic datasets (each having different number of users, features and samples). Numbers in bold indicate the highest mean AUC score for each dataset, while undelined numbers indicate cases where EBMs' score is close to the highest one.}
\scalebox{0.74}{
\begin{tabular}{ccc|cccccc}
\hline
\#Users & \#Feat   & \#Samples & EBM    & XGBoost       &   Grad        & RF       & SVM    & LogReg     \\
\hline
20 & 25 &  50 & 0.715 (0.149) & \textbf{0.747 (0.145)} & 0.699 (0.179) & 0.734 (0.168) & 0.638 (0.185) & 0.700 (0.149) \\
 20 & 25 & 100 & \textbf{0.736 (0.142)} & 0.707 (0.127) & 0.706 (0.132) & 0.735 (0.130) & 0.664 (0.130) & 0.702 (0.087) \\
 20 & 25 & 300 & \textbf{0.695 (0.154)} & 0.663 (0.148) & 0.678 (0.133) & 0.691 (0.147) & 0.684 (0.163) & 0.667 (0.157) \\
 20 & 60 & 100 & \underline{0.757 (0.147)} & \textbf{0.762 (0.181)} & 0.745 (0.153) & 0.760 (0.142) & 0.620 (0.147) & 0.634 (0.143) \\
 20 & 60 & 300 & \textbf{0.761 (0.127)} & 0.752 (0.121) & 0.749 (0.107) & 0.747 (0.127) & 0.672 (0.105) & 0.685 (0.113) \\
 50 & 25 &  50 & \textbf{0.736 (0.170)} & 0.722 (0.170) & 0.668 (0.157) & 0.711 (0.155) & 0.634 (0.188) & 0.657 (0.173) \\
 50 & 25 & 100 & 0.718 (0.128) & 0.718 (0.133) & 0.706 (0.128) & \textbf{0.726 (0.121)} & 0.655 (0.145) & 0.690 (0.132) \\
 50 & 25 & 300 & \underline{0.750 (0.111)} & 0.739 (0.108) & 0.741 (0.107) & \textbf{0.751 (0.111)} & 0.739 (0.123) & 0.744 (0.121) \\
 50 & 60 & 100 & \underline{0.680 (0.154)} & \textbf{0.684 (0.148)} & 0.675 (0.136) & 0.667 (0.148) & 0.558 (0.150) & 0.603 (0.136) \\
 50 & 60 & 300 & \textbf{0.764 (0.101)} & 0.755 (0.105) & 0.749 (0.103) & 0.757 (0.101) & 0.685 (0.101) & 0.701 (0.102) \\
100 & 25 &  50 & 0.688 (0.179) & 0.685 (0.158) & 0.670 (0.172) & \textbf{0.695 (0.148)} & 0.572 (0.193) & 0.629 (0.177) \\
100 & 25 & 100 & 0.675 (0.147) & 0.676 (0.144) & 0.671 (0.144) & \textbf{0.690 (0.147)} & 0.613 (0.133) & 0.618 (0.131) \\
100 & 25 & 300 & \underline{0.751 (0.110)} & 0.742 (0.101) & 0.744 (0.104) & \textbf{0.757 (0.109)} & 0.748 (0.109) & 0.748 (0.110) \\
100 & 60 & 100 & \textbf{0.737 (0.131)} & 0.711 (0.134) & 0.718 (0.122) & 0.696 (0.122) & 0.600 (0.131) & 0.634 (0.122) \\
100 & 60 & 300 & \textbf{0.722 (0.131)} & 0.709 (0.128) & 0.710 (0.117) & 0.710 (0.126) & 0.665 (0.091) & 0.668 (0.112) \\
\hline
\end{tabular}
}
\label{tab:syn_idio}
\end{table}

\subsubsection{Nomothetic Approach}
Subsequently, personalized EBMs are evaluated in comparison to the two nomothetic approaches described in Section \ref{method}, the using all data EBMs (EBM\_all) and knowledge distillation (KD) method. In case of knowledge distillation, different values for the temperature parameter are evaluated, ranging from 1 to 100. After applying all examined methods, the mean and standard deviation values of the produced AUC scores for each synthetic dataset are presented in Table \ref{tab:syn_nomo}.

\begin{table}[t]
\caption{Performance of the two nomothetic methods (EBM\_all and KD): the mean and standard deviation of the AUC scores are given for all synthetic datasets (each having different number of users, features and samples). Numbers in bold indicate the highest mean AUC score for each dataset, while undelined numbers indicate cases where distillation outperforms personalized EBMs.}
\scalebox{0.845}{
\begin{tabular}{ccc|ccccc}
\hline
\#User & \#Feat   & \#Samples & EBM                   & EBM\_all       & KD ($T = 1$)       & KD ($T = 5$) &      KD ($T = 100$) \\
\hline
20 & 25 &  50 & 0.715 (0.149) & \textbf{0.804 (0.151)} & 0.753 (0.178) & 0.768 (0.185) & \underline{0.776 (0.178)} \\
 20 & 25 & 100 & 0.736 (0.142) & \textbf{0.758 (0.162)} & 0.739 (0.134) & 0.735 (0.139) & \underline{0.753 (0.148)} \\
 20 & 25 & 300 & 0.695 (0.154) & 0.691 (0.172) & \textbf{0.698 (0.167)} & 0.694 (0.171) & 0.690 (0.179) \\
 20 & 60 & 100 & 0.757 (0.147) & \textbf{0.813 (0.111)} & 0.786 (0.092) & 0.779 (0.096) & \underline{0.795 (0.097)} \\
 20 & 60 & 300 & 0.761 (0.127) & \textbf{0.762 (0.119)} & 0.757 (0.111) & 0.756 (0.113) & \textbf{0.762 (0.119)} \\
 50 & 25 &  50 & 0.736 (0.170) & \textbf{
 0.756 (0.183)} & 0.707 (0.169) & 0.719 (0.170) & 0.731 (0.166) \\
 50 & 25 & 100 & 0.718 (0.128) & \textbf{0.747 (0.146)} & 0.713 (0.162) & 0.720 (0.164) & \underline{0.733 (0.160)} \\
 50 & 25 & 300 & 0.750 (0.111) & \textbf{0.773 (0.133)} & 0.762 (0.134) & \underline{0.769 (0.135)} & \underline{0.769 (0.135)} \\
 50 & 60 & 100 & 0.680 (0.154) & \textbf{0.735 (0.140)} & 0.689 (0.144) & 0.686 (0.147) & \underline{0.700 (0.151)} \\
 50 & 60 & 300 & 0.764 (0.101) & \textbf{0.783 (0.120)} & 0.751 (0.119) & 0.755 (0.122) & \underline{0.766 (0.123)} \\
100 & 25 &  50 & 0.688 (0.179) & \textbf{0.767 (0.175)} & 0.720 (0.167) & 0.725 (0.171) & \underline{0.736 (0.166)} \\
100 & 25 & 100 & 0.675 (0.147) & 0.723 (0.144) & 0.719 (0.138) & 0.720 (0.135) & \textbf{0.726 (0.141)} \\
100 & 25 & 300 & 0.751 (0.110) & \textbf{0.769 (0.121)} & \underline{0.767 (0.120)} & 0.765 (0.119) & 0.764 (0.121) \\
100 & 60 & 100 & 0.737 (0.131) & \textbf{0.761 (0.140)} & 0.712 (0.150) & 0.721 (0.147) & \underline{0.738 (0.148)} \\
100 & 60 & 300 & 0.722 (0.131) & \textbf{0.736 (0.142)} & 0.724 (0.133) & 0.720 (0.132) & \underline{0.729 (0.139)} \\
\hline
\end{tabular}
}
\label{tab:syn_nomo}
\end{table}

In the majority of the examined datasets, it is apparent that using personalized EBMs leads to worse performance than when either of the nomothetic methods is applied. More specifically, EBM\_all gives the best results compared to the distillation method in all but three datasets, whereas in one of these, both methods achieved the same score.
It is also interesting to mention that their difference, in terms of the mean AUC score, is quite big in some datasets. This is the case in datasets with a small number of samples, such as when characteristics (\{users, features, samples\}) are \{20, 25, 50\}, \{50, 25, 50\}, \{100, 25, 50\}, \{50, 60, 100\} and \{100, 60, 100\}. Therefore, it is important to highlight that collecting sufficient data from each user can benefit the knowledge distillation process. 

\subsection{Dataset: Drink}
\subsubsection{Idiographic Approach}

First, the total number of 24 individuals is analyzed using a personalized approach. After applying all different ML models, the results of the personalized predictive models on the testing sets indicated that the produced results highly vary across individuals. For instance, some individuals had quite high AUC results, whereas others’ results were at chance level.

\begin{figure}[!ht]
\centering
\begin{subfigure}[b]{0.555\textwidth}
     \centering
     \includegraphics[width=\textwidth]{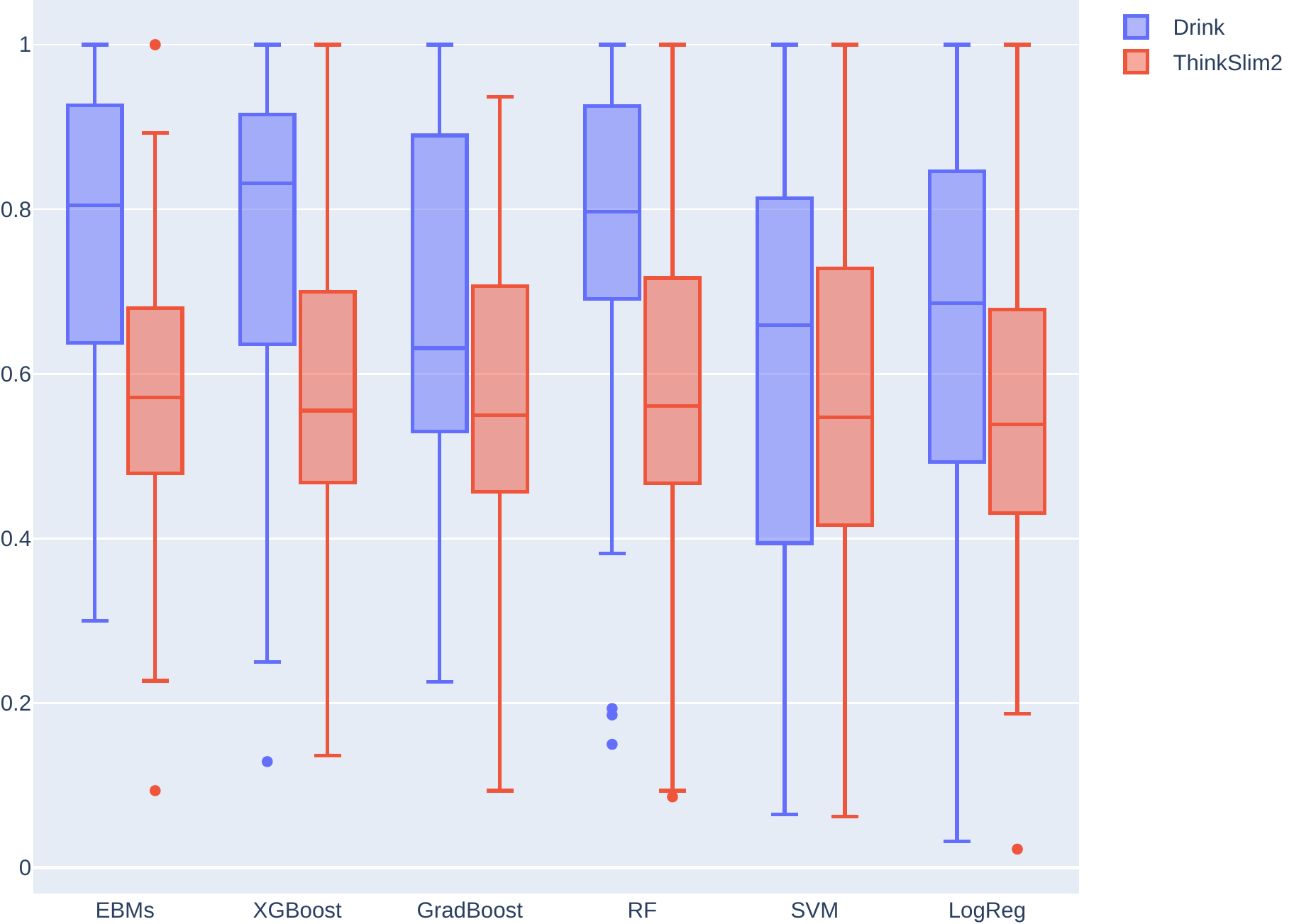}
     \caption{}
     \label{fig:a}
\end{subfigure}%
\begin{subfigure}[b]{0.395\textwidth}
     \centering
     \includegraphics[width=\textwidth]{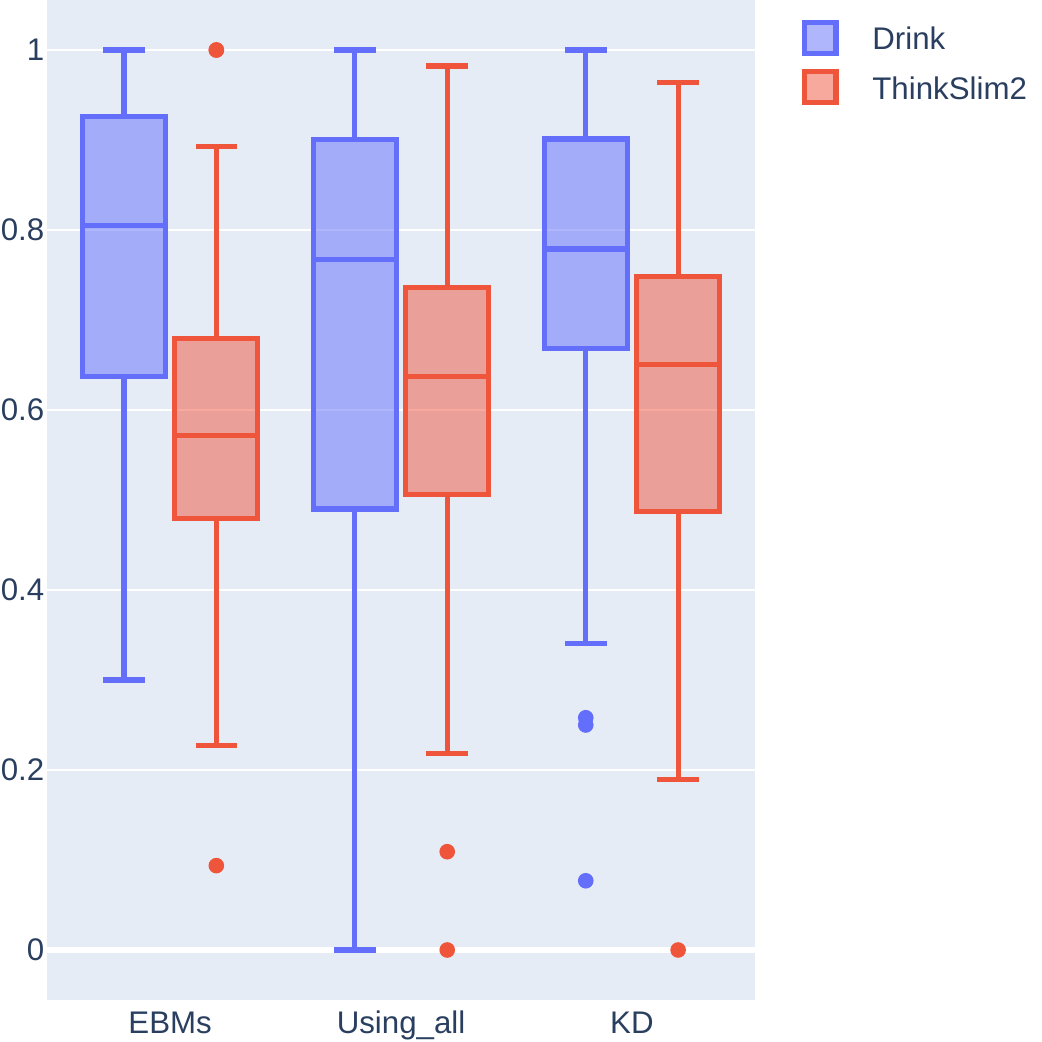}
     \caption{}
     \label{fig:b}
\end{subfigure}
\caption{a: AUC performance of all non-linear and linear models b: Comparing the performance of personalized EBMs to the two nomothetic approaches (EBM\_all and KD)}
\end{figure}

To compare the different ML models, we show some of the statistical properties of all AUC scores, using the box and whisker plots in Fig.~\ref{fig:a}. In this figure, we present the performance of EBMs compared to the full-complexity ML models as well as the performance of non-linear models compared to the traditionally-used linear ones. Regarding the first comparison, AUC's distribution for EBMs is comparable to the ones of the other non-linear models. Apart from RF, which shows a slightly better overall performance, all statistical properties of the EBMs' scores reached higher values than the other three models. The median value of EBM AUC score is around $0.81$, only a bit lower than XGBoost ($0.83$).
It can also be noticed that the minimun value of EBM performance was the highest among ML models, indicating a smaller variation among individuals in the case of EBMs. 

Regarding the second comparison, a distinction between the linear and non-linear models is clearly visible. All statistical properties of the AUC scores are lower in the case of linear models. These findings highlight the ability of non-linear ML models to enhance the predictive performance of the traditionally-applied linear ones.

\subsubsection{Nomothetic Approach}
In the nomothetic approach, data from all individuals are pooled into one dataset and modeled collectively by one EBM (EBM\_all), or further exploited in a personalized way (KD). To facilitate comparison, box and whisker plots are utilized and presented (as before) in Fig.~\ref{fig:b}.

Using a nomothetic approach, the AUC distribution of the KD method is improved compared to that of personalized EBMs. This shows more consistent performance scores across individuals, apart from 4 outliers. Regarding the EBM\_all method, its AUC distribution is more spread, meaning that the 25th percentile and minimum values are lower compared to personalized EBMs and KD. However, the upper half of its distribution is comparable to the respective part of the distributions obtained through the other cases. Subsequently, by comparing the median values of both approaches, we see that there is a slight distinction between them, where personalized EBMs reach the level of $0.80$, whereas around $0.76$ and $0.79$ for the EBM\_all and KD methods, respectively. In contrast to the results on synthetic datasets, we see that in a more realistic dataset, the knowledge distillation method can lead to improved results compared to EBM\_all.

\subsection{Dataset: ThinkSlim2}
\subsubsection{Idiographic Approach}

Similar to the previous dataset, the performance of $57$ personalized predictive models is first evaluated. As the produced results highly varied across individuals, their performance is also here assessed through box and whisker plots. Fig.~\ref{fig:a} presents the AUC scores of all different ML methods. According to AUC scores, all models’ distributions are comparable to each other, having a quite large range. All methods show similar poor performance, achieving a low median value around $0.57$ in the case of non-linear models, whereas around $0.54$ for the linear ones. That could be due to the more complex and challenging structure of this dataset, containing a larger number of individuals as well as features, but not more data samples compared to the previous dataset. Another interesting aspect in this experiment is that some AUC values are very close to zero (for all setups). This means that probabilities produced by all models for these individuals lead to a flipped prediction label for almost all testing points.

\subsubsection{Nomothetic Approach}
Finally, personalized EBMs were compared to the two nomothetic approaches, EBM\_all and KD. The results of all methods, in terms of AUC scores, are presented in Fig.~\ref{fig:b}. The median as well as the 25th and 75th percentile values are similar for both KD and EBM\_all, and also increased compared to the respective values of the personalized EBMs. The mean relative AUC increase of KD and EBM\_all compared to EBMs are at 17\% and 14\%, respectively. It is also worth mentioning that there is one individual having an AUC score equal to $0$. This means that the probabilities produced by both EBM\_all and KD methods for this individual do not map the class labels correctly, maybe because they are different than the rest of the population. In challenging problems, like the one represented by the ThinkSlim2 dataset, where personalized non-linear models do not perform well, both nomothetic approaches are likely to achieve a slightly improved performance.

\section{Challenges of modelling EMA data}
Studying the aforementioned two real-world datasets and noticting their varying results across individuals shows the importance of collecting good quality EMA data. Because of the complex nature of psychological behavior, its representation on a dataset can be quite challenging. EMA data collection is a difficult task, trying to capture multiple observations on subjective variables during an intensive period. Thus, it may contain unclear and arbitrary responses as well as missing values. 

Missing data is a significant problem of real-world EMA datasets that cannot be controlled during a study. Even though several individuals initially participate in a study, some may not produce enough data for analysis (especially if one needs to take into account the temporal nature of the data). The number of data points that is sufficient depends on the total compliance of each individual during the whole data collection period and also per day.
The most common approach to deal with missing data is to delete them while keeping only the complete sets of data. However, this method relies on the assumption that the missing observations are missing at random (MAR) or completely at random (MCAR), which possibly is not always the case.

\section{Conclusion}
This research work highlights the importance of exploiting the wealth of EMA data through more advanced ML models compared to linear ones. Non-linear vs. linear and idiographic vs. nomothetic approaches were investigated for classifying a target variable at a next time-point on different datasets. 

The results showed great consistency for the idiographic approach, showing that non-linear models yield an enhanced performance on both synthetic and real-world data. Subsequently, regarding the nomothetic approaches, no clear trends were observed in the results of all datasets. Although the EBM\_all method appears to perform best for synthetic datasets, that is not the case for the real-world datasets. Overall, the proposed knowledge distillation method could be recognized as the most beneficial to improve performance of personalized models. However, the differences in both idiographic and nomothetic approaches were not found statistically significant. As a future step, further experiments are needed on more (and larger) datasets for evaluating the examined approaches.

\subsection*{Acknowledgements}
This study is part of the project ``New Science of Mental Disorders"(www.nsmd.eu), supported by the Dutch Research Council and the Dutch Ministry of Education, Culture and Science (NWO gravitation grant number 024.004.016).  

%
%
%
\bibliographystyle{splncs04}
\bibliography{Bibliography}

\end{document}